\documentclass{article}
\usepackage{spconf,amsmath,graphicx}
\usepackage{arabtex}
\usepackage[ruled,vlined]{algorithm2e}
\usepackage[utf8]{inputenc}
\usepackage{utf8}
\usepackage[colorinlistoftodos]{todonotes}

\title{Creating Speech-to-Speech Corpus from Dubbed Series}
\name{Massa Baali$^1$, Wassim El-Hajj$^1$, Ahmed Ali$^2$}
\address{
  $^1$American University of Beirut, Computer Science Department, Beirut, Lebanon\\
  $^2$Qatar Computing Research Institute Doha, HBKU, Doha, Qatar}
\begin{document}
\ninept
\maketitle
\begin{abstract}
Dubbed series are gaining a lot of popularity in recent years with strong support from major media service providers. Such popularity is fueled by studies that showed that dubbed versions of TV shows are more popular than their subtitled equivalents. We propose an unsupervised approach to construct speech-to-speech corpus, aligned on short segment levels, to produce a parallel speech corpus in the source- and target- languages. Our methodology exploits video frames, speech recognition, machine translation, and noisy frames removal algorithms to match segments in both languages. To verify the performance of the proposed method, we apply it on long and short dubbed clips. Out of $36$ hours TR-AR dubbed series, our pipeline was able to generate $17$ hours of paired segments, which is about $47$\% of the corpus. We applied our method on another language pair, EN-AR, to ensure it is robust enough and not tuned for a specific language or a specific corpus. 
Regardless of the language pairs, the accuracy of the paired segments was around 70\% when evaluated using human subjective evaluation. The corpus will be freely available for the research community.
\end{abstract}
\section{Introduction} \label{sec:intro} 

Clean and large parallel speech corpora constitute a major building block in developing speech-to-speech translation systems \cite{jia1904direct}. However, building such a big corpora is a very costly and lengthy process that normally relies on manual labor. Many researchers focused on creating parallel corpora addressing manual annotations or text inputs, such as subtitle \cite{feng2017building, kajiwara2016building, tiedemann2007improved, cuvrin2004building, post2012constructing, avramidis2012richly, cattoni2021must}. Unlike the text corpora, speech corpora are very limited. Thus, the solutions that are proposed for building parallel speech resources are very few. Despite the increasing attention speech-to-speech research has received in the last few years \cite{jia1904direct,kano2018structured, kano2021transformer}, researchers in the field are lacking publicly available corpora to train data-hungry deep space systems. Given that there are very few language pairs that are covered, there is an urgent need for research and investigations in this domain.

Researchers in \cite{oktem2017automatic} extracted parallel speech corpora based on any language pairs from dubbed movies, in which some corresponding prosodic parameters are extracted, making it useful for other research fields, such as large comparative linguistic and prosodic studies. The authors, however, applied their version on a very small corpus ($80$ minutes), not to mention that they relied on subtitles. 
Tsiartas et al. \cite{tsiartas2011bilingual}  explored a method based on machine learning for automatically extracting bilingual audio subtitle pairs from movies. They used raw movie data and defined the long term spectral distance, subtitles time distance, and subtitle time-stamps to segment the bilingual speech regions. This work, however, followed a supervised approach. It also depended on subtitles that are manually tagged which is time consuming. 

In this work, we propose an unsupervised approach that takes a dubbed series input and produces a Speech-to-Speech Corpus in the respective languages of the dubbed series. We address the following major challenges: (\textit{i}) removing the noisy voice segments, such as commercials, from each dubbed version to produce dubbed versions that have almost the same duration; (\textit{ii}) reducing the effect of the background noise or music when matching the speech segments; and  (\textit{iii}) finally, carefully optimizing a set of heuristic rules for segments' matching, where hyperparameters can be tuned to optimize the tradeoff between quality and corpus size. The proposed approach is language independent in the sense that it accepts dubbed clips in any two languages and generate the parallel corpus in these languages. Hence, to demonstrate the efficiency of the proposed approach and its language independence, we test our pipeline on long TR-AR series and short EN-AR series. We picked TR-AR series following studies that showed that $85$ million Arab viewers watch Syrian-dubbed Turkish series \cite{buccianti2010dubbed}. Additionally, Arabic is a morphologically complex language with high degree of affixation and re-ordering when it is compared with English, making it the perfect language for such studies. 

In our study, the original input of the Turkish series was $51$ hours that correspond to $54$ hours of the dubbed Arabic version (in the Arabic version there were a lot of commercials). After cleaning the versions from the commercials, noise segments, and unrecognized segments, $36$ hours of each dubbed version were produced. After applying the segments' matching rules and tuning the hyperparameters to gain good quality parallel corpus, $17$ hours of parallel TR-AR speech corpus was produced. The evaluation that relied on random samples annotated by bilingual speakers showed a $70$\% overall accuracy. To demonstrate language independence, we also experiment with different languages, mainly EN-AR clips, where similar results were produced.  

Our proposed system builds a parallel speech corpora starting by analyzing the video modality. Table \ref{tab:comparison} benchmarks our approach compared to previous studies. 

Up to the best of our knowledge, this is the first study to use visual information as well as acoustic information to automate the process of creating speech to speech corpus.

\begin{table*}[hbt!]
\centering
\caption{Benchmarking our studies with previous work.}
\scalebox{0.9}{
\begin{tabular}{llllllll}
\hline
 \textbf{Paper} & \textbf{Unsupervised} & \textbf{Language} & \textbf{Result} &  \textbf {Speech} & \textbf{Bilingual} 
&  \textbf{Dub}
\\
 \hline

Ours&  Y&  Any&  \vtop{\hbox{\strut 70\% overall}\hbox{\strut  accuracy}} &  Y & Y&  Y \\ \hline

\cite{oktem2017automatic}&  Y&  Any&  - &  Y & Y &  Y \\ \hline

\cite{tsiartas2011bilingual}&  N&  English-French& \vtop{\hbox{\strut  LTSD 41.39\% }\hbox{\strut  Subtitle 37.88\%}} &  Y & Y &  Y \\ \hline

\cite{avramidis2012richly}&  N&  English-Czech,German,Spanish&  
BLEU 18.95 &  N & Y &  N \\ \hline

\cite{post2012constructing}&  N& \vtop{\hbox{\strut English and 6 languages}\hbox{\strut  from Indian sub-contents}} &  
\vtop{\hbox{\strut 5 votes cast on 65\%}\hbox{\strut  of sentences}} &  N & Y&   N \\ \hline

\cite{cuvrin2004building}&  N& Czech-English & - &  N & Y&  N \\ \hline

\cite{tiedemann2007improved}&  N& 29 languages & \vtop{\hbox{\strut 85\% correct }\hbox{\strut  alignment}} &  N & Y&  N \\ \hline

\cite{feng2017building}&  N& Chinese-English & \vtop{\hbox{\strut 90\% annotation}\hbox{\strut  agreement}} &  N & Y &  N \\ \hline
 \cite{kajiwara2016building}&  Y& English & BLEU 26.3&  N & N&   N \\ \hline
\end{tabular}
}
\label{tab:comparison}
\end{table*}

\section{Parallel Speech Corpus Construction} \label{sec:data}
\begin{figure}
\caption{Pipeline for the Parallel Speech Corpus  Construction: data collection, voice activity detection (VAD), automatic speech recognition (ASR), machine translation (MT).}
    \centering
    \begin{minipage}[b]{0.70\linewidth}
    \centering
        \centerline{\includegraphics[width=9cm]{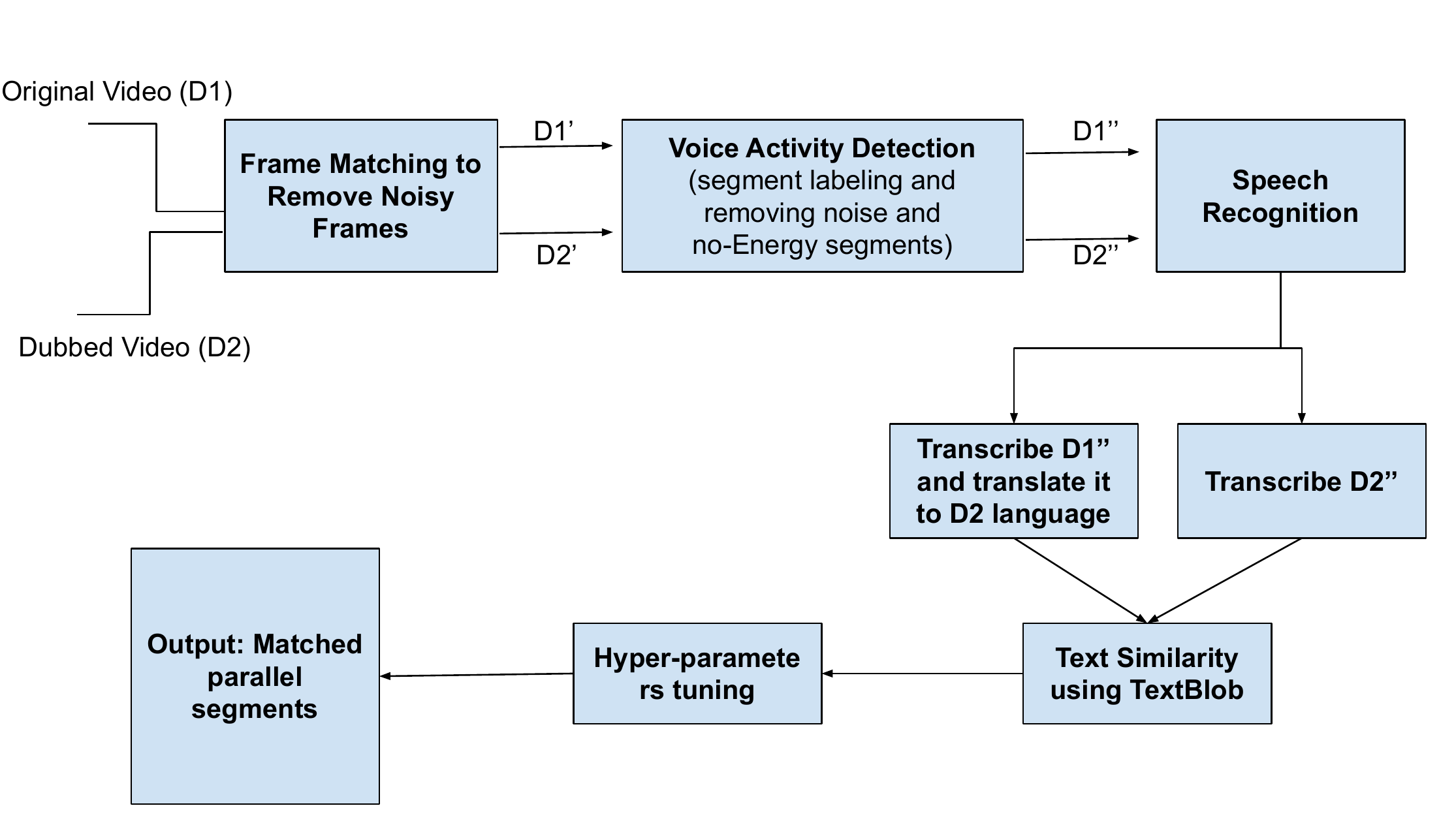}}
    \end{minipage}
%
    
    \label{pipeline}
\end{figure}

The process of creating 
parallel speech corpus from dubbed series goes through the following steps: (\textit{i}) data collection, (\textit{ii}) voice activity detection, (\textit{iii}) segmentation, (\textit{iv}) automatic speech recognition, (\textit{v}) textual machine translation, and finally (\textit{vi}) segments matching. Figure \ref{pipeline} illustrates an overview of this pipeline. Each step of the pipeline is explained using examples from the TR-AR dubbed series. The same pipeline could be used for any dubbed series or any language pairs without losing generality. 

\subsection{Data Collection and Video Matching} \label{seubsec:data_collection}
In this phase, we aim to download and clean a dubbed video in any language pair. In some cases, you might be able to get hold of dubbed videos with equal duration and without any additional noise segments such as commercials; in which case, the videos are ready to be processed. In other cases, one or both of the dubbed versions might include noise segments that should be removed, in order to end up with a clean version of the dubbed videos. Algorithm \ref{alg1} takes as input the dubbed versions and cleans them up. The first step in the algorithm is to convert every dubbed video into frames, $30$ frames per second.  At time t, we pick the frame in D$1$ and compare it with $500$ consecutive frames in D$2$ (also starting at t) i.e. we are searching whether the frame in D$1$ at time t exists in a $16$ seconds video segment in D$2$ starting at time t. Every two frames are compared using skimage \cite{van2014scikit} to compute the mean structural similarity index between two images. The values returned by skimage are in the range of [$0$,$1$], where a value close to $1$ indicates high similarity between the two images. In case two frames returned a similarity greater than $0.75$, we assume that these two images are similar, and thus the frame we are investigating is a valid frame, and not a noise frame. This process is repeated  every frame in D$1$. A frame in D$1$ with low similarity in the corresponding consecutive frames in D$2$ is removed.  Once done, we run the same algorithm starting with the other dubbed version. For instance, all frames in D$1$ that correspond to images from commercial segment will be removed, since no matching frames exist for them in D$2$.

\begin{algorithm}[ht]
\SetAlgoLined
D$1$: Dubbed version $1$\\ 
D$2$: Dubbed version $2$\\
\KwResult{Clean and matched dubbed videos}
  Convert D1 and D2 into frames, 30 frames per second;\\
  \While{there are more frames f in D1}{
  Extract frame f at time t from D1\\
  t' = t \\
  \While{i \textless 500}{
    Extract frame f' at time t' from D2 \\
    r = skimage(f, f') \\
    \eIf{r $\geq$ 0.75} {break;} 
  {t'=t'+1 \\
    i=i+1}

  }

  \If{i==499}{
    remove frame f at time t from D1}
  t = t + 1 (move to analyze the next frame in D1)
 }
 \caption{Proposed procedure to remove noise frames from the dubbed videos}
 \label{alg1}
\end{algorithm}

To apply the algorithm on real data, we downloaded three Turkish (TR) - Arabic (AR) dubbed series from YouTube. The three Turkish series have total durations of $12$, $9$, and $30$ hours, respectively. The corresponding Arabic series have total durations of $13$, $10$, and $31$ hours, respectively. When checking the series manually, we noticed that the Arabic version contained a lot of commercials, the fact that explains the extra duration of the Arabic series when compared to the Turkish series. After running Algorithm \ref{alg1}, we ended up with a matching duration of both series equivalent to $12$, $9$, and $30$ hours respectively. Moving forward, we operate on all series as one big video; i.e., the new input is a TR-AR dubbed video pair composed of $51$ hours each. 

\subsection{Voice Activity Detection} \label{subsec:VAD}
Algorithm \ref{alg1} produced matching dubbed videos without noise frames, such as commercials. 
We now extract the audio from the dubbed video samples at 16Khz. Then, we process the audio files of each dubbed videos. We investigate the impact of segmentation using voice activity detection (VAD). We use the inaSpeechSegmenter pre-trained model \cite{doukhan2018open} that extract meta-data from each audio file; gender, noise, music and noEnergy (silence).

\begin{figure}[h!]
    \centering
    \begin{minipage}[b]{0.70\linewidth}
  \centering
  \centerline{\includegraphics[width=7cm]{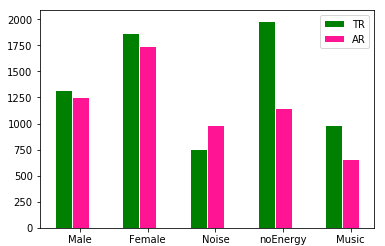}}
\end{minipage}
\caption{Frequency of TR-AR labeled segments, \textit{y-axis} is the number of segments.}

    \label{dubbedVideosLabels}
\end{figure}

By applying this phase on the TR-AR dubbed series. Figure \ref{dubbedVideosLabels} shows the distribution of the five labels in the Turkish Dubbed version and the Arabic Dubbed version. The frequency of most labels is similar with the exception of noEnergy. We attribute this to the Arabic dubbing, where the actors ignore the amount of silence in the original video while abiding by the scene time. 

\subsection{Speech Segments Matching} \label{subsec:text_per_chunck}
\begin{figure}[h!]
    \centering
    \begin{minipage}[b]{0.70\linewidth}
    \centering
        \centerline{\includegraphics[width=10cm]{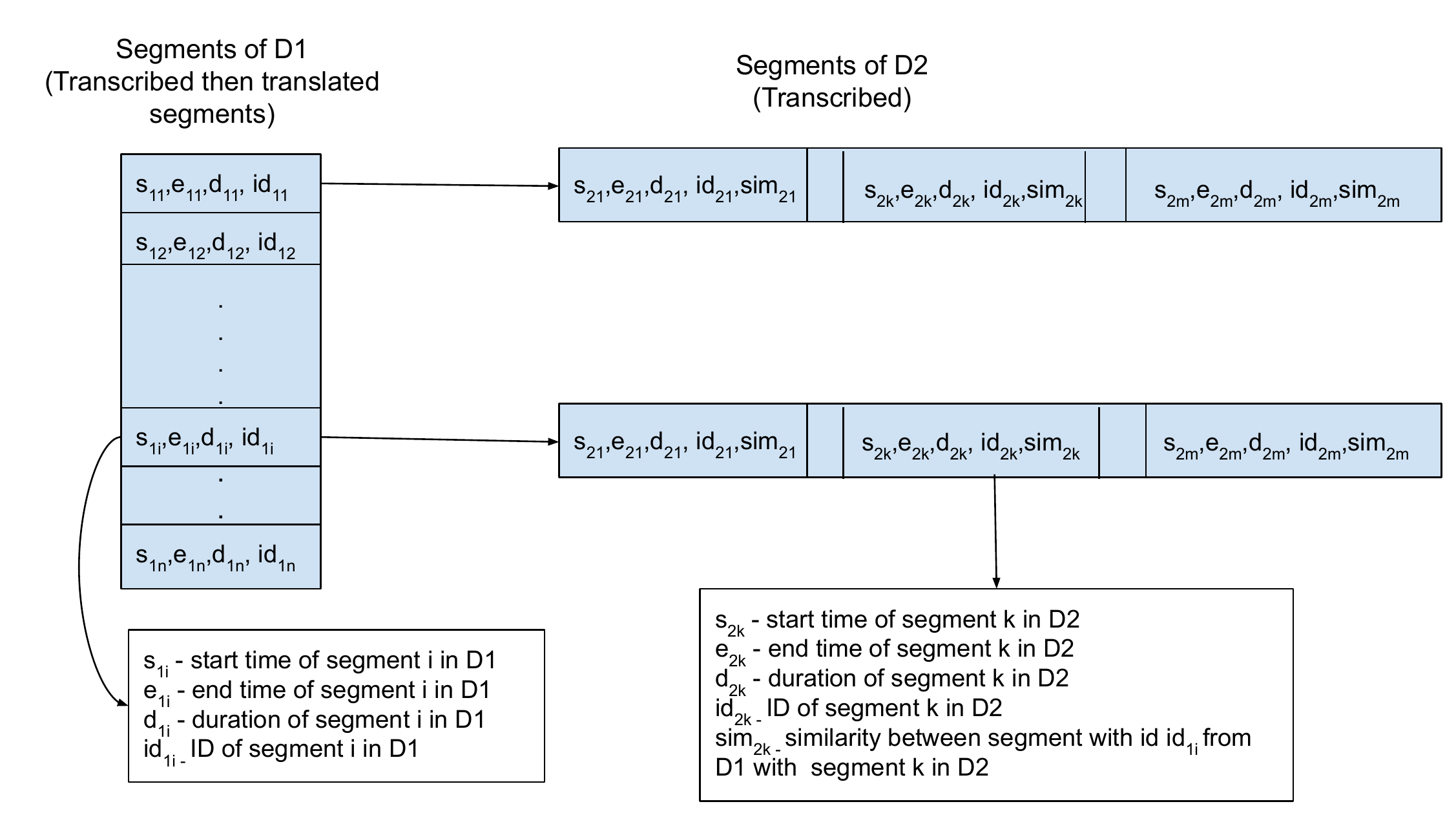}}
    \end{minipage}
    \caption{Similarity matching between a segment in D1 (first dubbed version) and all segments in D2 (second dubbed version)}

    \label{similarity}
\end{figure}

The VAD $2$ produced speech segments that are labeled as female, male, and music. Every segment has a start time, end time, and duration. In this phase, We transcribe and translate each segment using Google API. It is worth noting that we found that transcription accuracy varies a lot as some audio sections can be very challenging to recognize. As a result of that, we obtain transcription for every segment. 
We then translate the segments' transcription of one dubbed version to the  language of the other dubbed version. For example, we translate the Turkish transcribed segments into Modern Standard Arabic (MSA). 
Any segment that was not recognized by the transcription API is removed. By applying this step on the TR-AR dubbed versions, we ended up with $36$ hours videos that are candidates for matching. Next, we calculate the similarity between every translated segment in one dubbed version (e.g. the Arabic text translated from the Turkish transcribed segments), and all the transcribed segments in the other version (e.g. the Arabic text transcribed from the Arabic segments). Since the pipeline comprises of speech recognition, machins translation as well as the nature of the dubbing is not verbatim, we decide not to use the standard word error rate evaluation metric and look for text similarity score between the source- and target- segments. 
We use TextBlob \cite{loria2018textblob}, a library which is based on gensim and Fasttext pretrained word2vec model to measure the distance using cosine similarity. 
Figure \ref{similarity} captures the data structure that is used to record the similarity between every translated segment in the first dubbed version and its corresponding transcribed version.

The next step is to match one or more segments in the first dubbed version (e.g. TR) to one or more segments in the second dubbed version (e.g. AR), with the following three possibilities: 
\begin{enumerate}
\item  Matching one segment in the first dubbed version to one segment in the second dubbed version.
\item  Matching one long segment in the first dubbed version to many segments in the second dubbed version.
\item Matching many segments in the first dubbed version to one long segment in the second dubbed version.
\end{enumerate}
To do the matching according to one of three mentioned possibilities, we devise the following set of matching rules based on the start time, end time, duration, label, and similarity score of each segment:

\begin{itemize}
  \item Rule 1: Difference between the segment start time in the first version and the segment start time in the second version, is below a certain threshold. 
  \item Rule 2: Difference between the segment duration in the first version and the segment duration in the second version, is below a certain threshold.
  \item Rule 3: The segments in both versions should have the same label. 
  \item Rule 4: The similarity score between two segments is above 0.5. 
  \item Rule 5: To combine multiple short segments in one version and match them with one long segment in another version, we adopt a sliding window approach that starts with one short segment abiding by Rule 1, then continues adding more consecutive short segments in the same version until the cumulative duration of the short segments abides by Rule 2, and the similarity score abides by Rule 4.  
\end{itemize}

\begin{table*}[h!]
\centering
\caption{{Results when running the pipeline on 36 hours of TR-AR dubbed series}}
\resizebox{18cm}{!}{

\begin{tabular}{lllllllll}
\hline
\textbf{Lang.} & \textbf{Input Duration} & \textbf{Input Segments} & 
\textbf{Dif Start Time} &  \textbf {Dif Dur} & \textbf{Output Segments} 
& \textbf{Output Duration}&\textbf{Avg Similarity} & \textbf{Percent}
\\ \hline
TR-AR& 36 hrs&28,800;27,678&\textless=3&\textless=2&9,998;9,887&14.3hrs&0.55 & 39\%\\ \hline
TR-AR& \textbf{36 hrs}&\textbf{28,800;27,678}&\textbf{\textless=9}&\textbf{\textless=8}&\textbf{11,581;10,060}&\textbf{17.6hrs}&\textbf{0.54} & \textbf{ 48\%}\\ \hline

\end{tabular}}
\label{rules}
\end{table*}

\begin{table*}[h!]
\centering
\caption{{Results when running the pipeline on 11 minutes and 1.2 hours of EN-AR dubbed clips}}
\resizebox{18cm}{!}{
\begin{tabular}{lllllllll}
\hline
\textbf{Lang.} & \textbf{Input Duration} & \textbf{Input Segments} & 
\textbf{Dif Start Time} &  \textbf {Dif Dur} & \textbf{Output Segments} 
& \textbf{Output Duration}&\textbf{Avg Similarity} & \textbf{Percent}
\\ \hline
EN-AR& 11 mins &61;112 &\textless=3&\textless=2&10;10&2 mins&0.56 & 18\%\\ \hline
EN-AR&\textbf{11 mins} &\textbf{61;112} &\textbf{\textless=9}&\textbf{\textless=8}&\textbf{18;18}&\textbf{4 mins}&\textbf{0.54} & \textbf{36\%}\\ \hline
EN-AR&1.2 hrs &1000;928 &\textless=3&\textless=2&400;400&15 mins&0.55 & 25\%\\ \hline
EN-AR&\textbf{1.2 hrs} &\textbf{1000;928} &\textbf{\textless=9}&\textbf{\textless=8}&\textbf{650;650}&\textbf{20 mins}&\textbf{0.54} & \textbf{33\%}\\ \hline
\end{tabular}}
\label{rulesENAR}
\end{table*}

We run multiple experiments tuning these hyper-parameters aiming to maximize the corpus size while maintaining a similarity score above $0.5$. 
We applied the algorithms and rules mentioned in this section to the TR-AR dubbed video. Table \ref{rules} highlights the application of the proposed methodology on dubbed videos. Every row in the table presents the original  duration (Input Duration) of the video and the corresponding number of extracted segments (Input Segments). Every row also contains values that represent the hyper-parameters, namely, \textit{Dif Start Time}  and \textit{Dif Dur} as presented in Rules 1 and 2. The last four entries in every row indicate the final number of parallel segments (Output Segments), their duration (Output Duration), their average similarity score (Avg Sim), and the percentage duration of parallel corpus produced.

As shown in table \ref{rules}, the difference-start-time threshold of $9$ seconds and the difference-duration threshold of $8$ seconds, produce a parallel corpus of duration $17.6$ hours; i.e. $48$\% of the cleaned dubbed videos can be transformed into a parallel corpus. When compared to the original dubbed TR-AR videos before cleaning ($51$ hours), we are able to produce a parallel speech corpus of duration $36$\% of the original videos. It is also worth noting that the rules above produce an appropriate average similarity score when varying the thresholds, while being able to extract a good percentage of parallel corpus duration. 

To evaluate whether the proposed approach generalizes well to other languages, we applied the same pipeline on EN-AR dubbed clips without any tuning. We chose two clips; short one, $11$ minutes long, and another longer clip, $1.2$ hours long. Using the same hyperparameters that gave the best results on the TR-AR dubbed series (difference-start-time = $9$, and difference-duration = $8$), the algorithms produced $4$ minutes and $20$ minutes of parallel speech segments respectively. On average, $35$\% of the original clips time. Table \ref{rulesENAR} captures the results. We can cautiously claim that the proposed pipeline produces around $35$\% of parallel speech duration with respect to the duration of the original dubbed input, regardless of the language.  We next evaluate the quality of the produced parallel speech corpus.


\begin{figure}[h!]
    \caption{Examples for Human Evaluation from TR-AR Corpus.}
    \scalebox{0.55}{
    \includegraphics{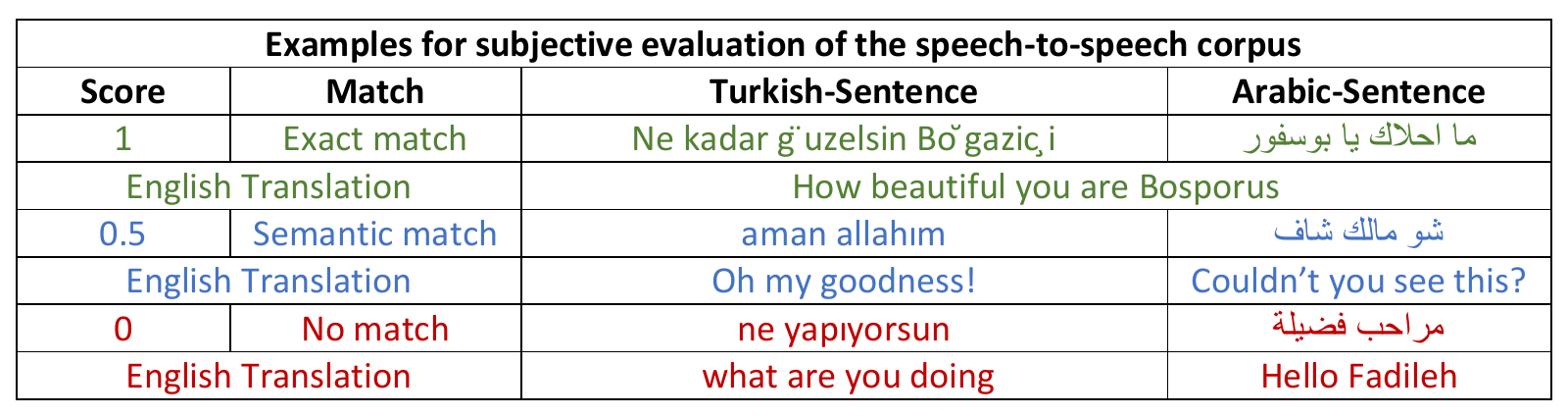}
    }
    \label{Dubbed_Example_cropped}
\end{figure}

\begin{table}[h]
\centering
\caption{{Label distribution for each score.}}
\resizebox{8cm}{!}{
\scalebox{0.8}{
\begin{tabular} {lllll} 
\hline
\textbf{Score} & \textbf{total}&\textbf{female} & \textbf{male}& \textbf{music}
\\ \hline
1	&512&265&154&93	\\ \hline
0.5	&163&88&53&22	\\ \hline
0	&325&173&102&50	\\ \hline
\end{tabular}}
}
\label{annotationSimilarity}
\end{table}

\section{Evaluation} \label{sec:eval}
Finally, we conduct subjective evaluation using mean opinion score MOS on the segments matching. We recruited two annotators and we asked them to rate the segments match on 3-point scale; 1 for exact match, 0.5 for semantic match and 0 for no match at all. To make sure that the annotation process is consistent, we asked the workers to annotate $100$ samples from each of the two corpora TR-AR and EN-AR. We used Cohen's kappa coefficient to measure the Inter-Annotator Agreement. The Kappa score was $0.79$ for the TR-AR segments, and $0.75$ for the EN-AR segments indicating substantial agreement.
For the TR-AR segments, a random sample of $1,000$ speech segments were given to the two bilingual annotators. The duration of the segments ranged from $2$ to $10$ seconds. Figure \ref{Dubbed_Example_cropped} shows example for scoring for the following three criteria:  (\textit{i}) score $1$ if the pairs should hold the same exact meaning; score (\textit{ii}) $0.5$ if the pairs are semantically matched, the video segments should be almost identical. It is worth noting, that semantic matching varies from country/language to another due to exaggeration applied by the dubbers in particular scenes to fit the culture of that country in terms of way of speaking. In the example shown in table \ref{Dubbed_Example_cropped}, it represents a scene for a person riding his bike and by accident he collided with a pedestrian. The expression was acted differently in each language but they stills hold the same meaning for that scene and referring to the same video section. (\textit{iii}) Finally, score $0$ means that the pair is not matching. Table \ref{annotationSimilarity} shows the agreement results, where $512$ ($52$\%) segment pairs were identical, $163$ ($17$\%) segment pairs are semantically similar, and $352$ ($31$\%) segments pairs were not similar. We can argue that the overall the similarity is around $70$\% given that $0.5$ is a correct match as far as the dubbed corpus is concerned.
We did the same exercises for the  EN-AR segments and we achieved around $70$\% segment match; both identical and semantic match.

\section{Conclusion} \label{sec:conclusion}
In this work, we introduced an unsupervised approach to creating parallel speech corpora from dubbed videos. Unlike existing approaches that are either supervised, or unsupervised but inefficient, the proposed approach can be tuned to produce large parallel speech corpora with high quality. For future work, we plan to study the effectiveness of the approach on dubbed clips in various languages.

\bibliographystyle{IEEEbib}
\bibliography{refs}

\end{document}